\documentclass[letterpaper]{article} 
\usepackage{aaai24}  
\usepackage{times}  
\usepackage{helvet}  
\usepackage{courier}  
\usepackage[hyphens]{url}  
\usepackage{graphicx} 
\usepackage{natbib}  
\usepackage{caption} 
\usepackage{algorithm}

\usepackage{algorithm,algpseudocode}
\usepackage{varwidth}
\usepackage{algorithmicx,algorithm}
\usepackage{wrapfig}
\usepackage{lipsum}
\usepackage{adjustbox}

\usepackage{amsmath}
\usepackage{amssymb}
\usepackage{mathtools}
\usepackage{amsthm}

\usepackage{microtype}
\usepackage{graphicx}
\usepackage{subcaption}
\usepackage{booktabs} 
\usepackage{multirow}
\usepackage{enumitem}
\usepackage{bm}
\usepackage{hyperref}
\hypersetup{
    colorlinks=true,
    linkcolor=red,
    citecolor=blue,
    filecolor=magenta,      
    urlcolor=blue,
linktocpage}
\usepackage[figuresright]{rotating}
\usepackage[dvipsnames]{xcolor}
\newcommand*\samethanks[1][\value{footnote}]{\footnotemark[#1]}
\newcommand{\blue}[1]{\textcolor{blue}{#1}}
\usepackage{newfloat}
\usepackage{listings}
\DeclareCaptionStyle{ruled}{labelfont=normalfont,labelsep=colon,strut=off} 
\floatstyle{ruled}
\newfloat{listing}{tb}{lst}{}
\floatname{listing}{Listing}
\title{A Perspective of Q-value Estimation on\\  Offline-to-Online Reinforcement Learning}
\author {
    Yinmin Zhang\textsuperscript{\rm 1, 2} \thanks{Equal contribution. Author ordering is determined by coin flip.}\quad
    Jie Liu\textsuperscript{\rm 2, 3} \samethanks \quad 
    Chuming Li\textsuperscript{\rm 1, 2} \samethanks \quad
    Yazhe Niu\textsuperscript{\rm 2}, \\
    Yaodong Yang\textsuperscript{\rm 4}\thanks{Corresponding author} \quad
    Yu Liu\textsuperscript{\rm 2}\samethanks \quad
    Wanli Ouyang\textsuperscript{\rm 2}
}
\affiliations {
    \textsuperscript{\rm 1} The University of Sydney, SenseTime Computer Vision Group, Australia\\
    \textsuperscript{\rm 2} Shanghai Artificial Intelligence Laboratory \quad
    \textsuperscript{\rm 3} CUHK MMLab \quad
    \textsuperscript{\rm 4} Institute for AI, Peking University \\
    yinmin.zhang@sydney.edu.au\quad
    jieliu@ie.cuhk.edu.hk\quad
    chli3951@uni.sydney.edu.au \\
    yaodong.yang@pku.edu.cn \quad
    liuyuisanai@gmail.com\quad
    ouyangwanli@pjlab.org.cn
}

\usepackage{bibentry}

\begin{document}

\maketitle

\begin{abstract}

Offline-to-online Reinforcement Learning (O2O RL) aims to improve the performance of offline pretrained policy using only a few online samples.
Built on offline RL algorithms, most O2O methods focus on the balance between RL objective and pessimism, or the utilization of offline and online samples. 
In this paper, from a novel perspective, we systematically study the challenges that remain in O2O RL and identify that the reason behind the slow improvement of the performance and the instability of online finetuning lies in the inaccurate Q-value estimation inherited from offline pretraining.
Specifically, we demonstrate that the estimation bias and the inaccurate rank of Q-value cause a misleading signal for the policy update, making the standard offline RL algorithms, such as CQL and TD3-BC, ineffective in the online finetuning.
Based on this observation, we address the problem of Q-value estimation by two techniques: (1) perturbed value update and (2)  increased frequency of Q-value updates. The first technique smooths out biased Q-value estimation with sharp peaks, preventing early-stage policy exploitation of sub-optimal actions. The second one alleviates the estimation bias inherited from offline pretraining by accelerating learning.
Extensive experiments on the MuJoco and Adroit environments demonstrate that the proposed method, named SO2, significantly alleviates Q-value estimation issues, and consistently improves the performance against the state-of-the-art methods by up to 83.1\%.
\end{abstract}

\section{Introduction}
In recent years, deep offline Reinforcement Learning (RL) \citep{mopo, rem, pbrl} has received increasing attention due to its potential for leveraging abundant logged data or expert knowledge (\textit{e.g.}, human demonstrations) to learn high-quality policies.
Analogous to trends in computer vision~\cite{mae,clip,foundmodel} and natural language processing~\citep{bert,gpt3}, where powerful models pretrained on large, diverse datasets generalize to task-specific data through finetuning, \textit{offline-to-online reinforcement learning} (O2O RL) aims to enhance the performance of a pretrained offline RL policy via finetuning on only a few online collected samples.
In O2O RL, \textit{the central challenge is to maximize the efficient utilization of additional online samples for further performance improvement.} 

\begin{figure*}[t]
  \centering
\resizebox{0.95\linewidth}{!}{

\subfloat[Learning curve]{
\begin{minipage}{0.33\textwidth}
    \includegraphics[width=1.0\linewidth]{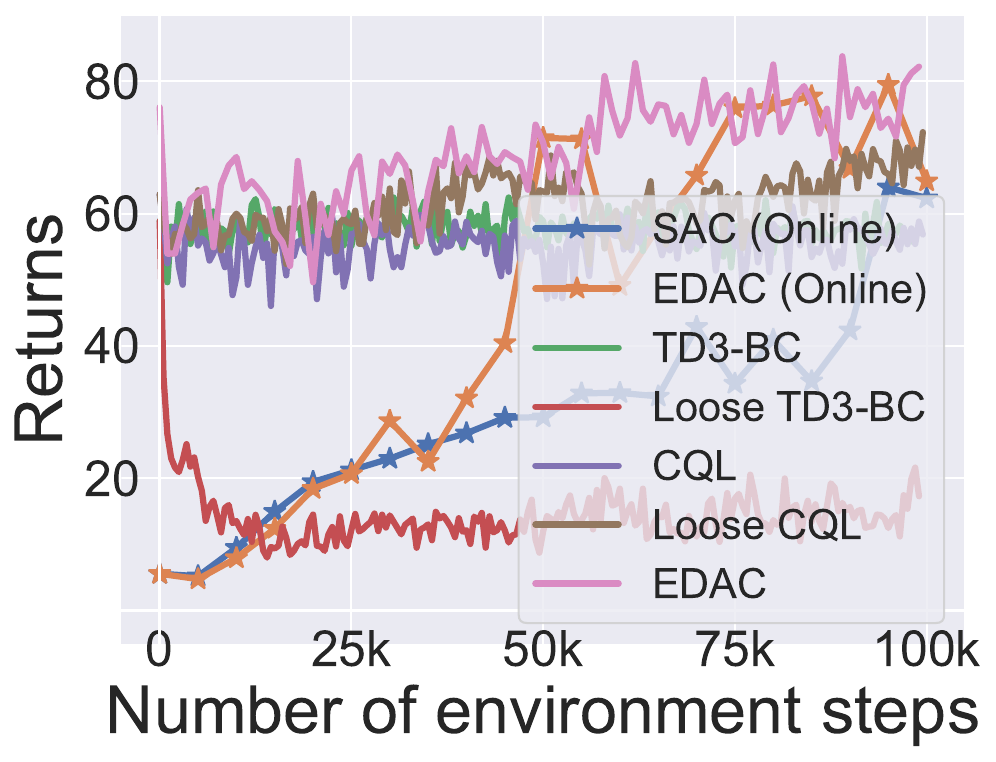}
    \label{fig:finetune_with_online_curve}
    \end{minipage}
    \vspace{-3ex}
}
\subfloat[Normalized difference of Q-value]{
\begin{minipage}{0.33\textwidth}
    \includegraphics[width=1.0\linewidth]{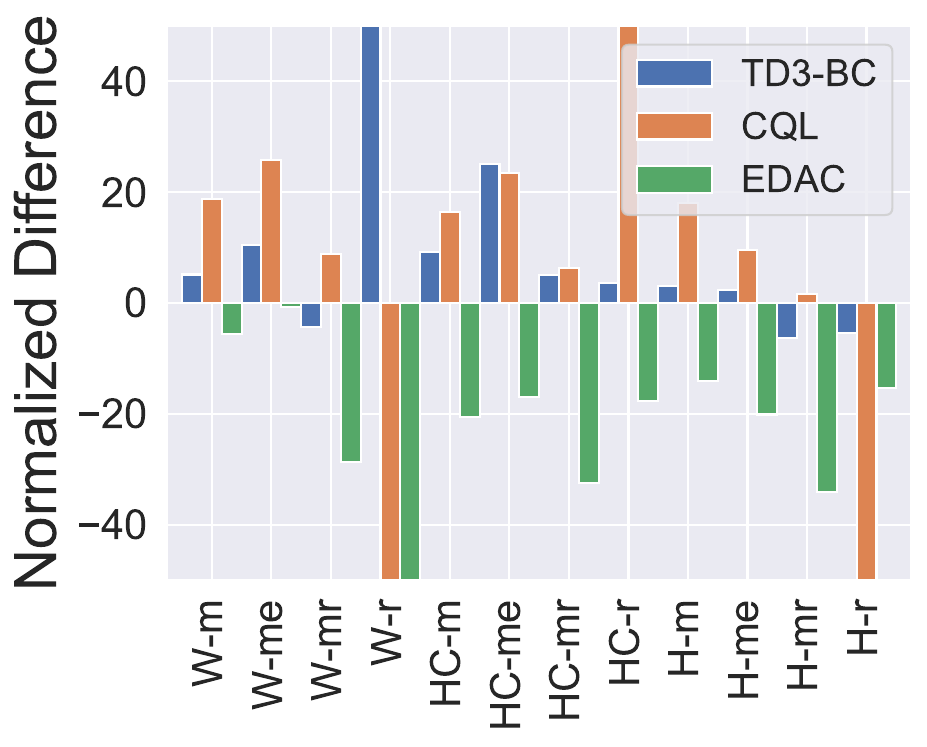}
    \label{fig:percent_diff}
    \end{minipage}
    \vspace{-3.9ex}
}
\subfloat[Kendall coefficient for Q-value]{
\begin{minipage}[t]{0.33\textwidth}
    \includegraphics[width=1.0\linewidth]{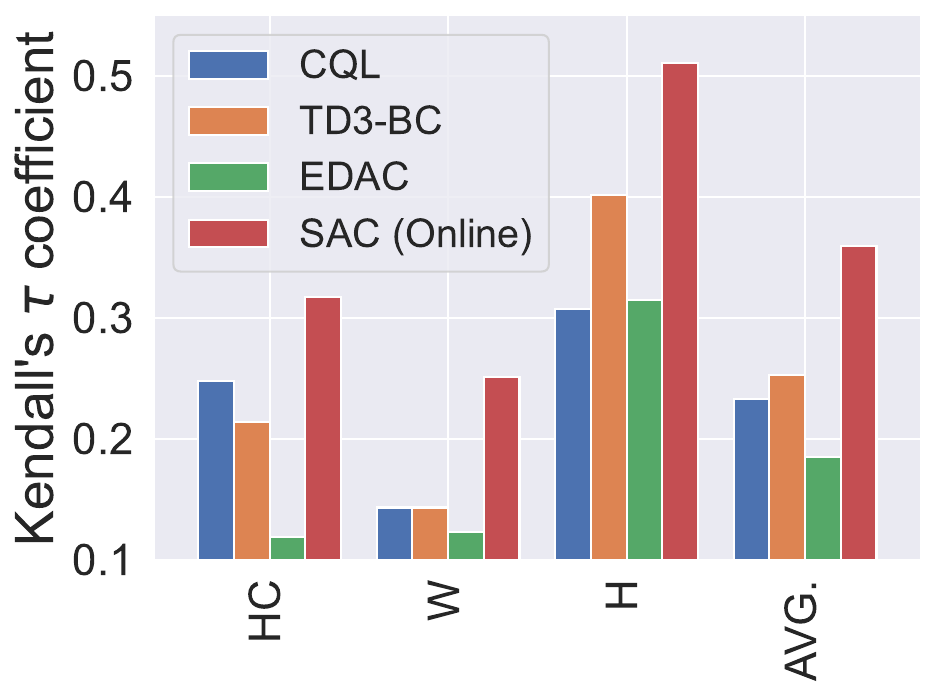}
    \label{fig:kendall}
    \end{minipage}
    \vspace{-2.4ex}
}
}
\caption{
\textbf{(a) Learning curves} on online finetuning.
  \textbf{(b) Normalized difference of the Q-value estimation} of offline RL algorithms, compared to an online RL baseline.
  A positive normalized difference means that the Q-value estimation of the offline RL algorithm tends to overestimate, compared to online RL, and vice versa. While offline algorithms deal with the out-of-distribution problem via a pessimistic perspective, TD3-BC and CQL usually overestimate the Q-value for the state-action pairs collected by the corresponding trained policy with noise.
  \textbf{(c) Kendall's $\tau$ coefficient}  between the Q-value estimation and the true Q-value on the same batch of state-action pairs. The lower coefficient shows that the rank of Q-value estimation from offline RL algorithms is severely worse than online RL algorithms, which leads to unstable training and slow improvement.
  HC = HalfCheetah, H = Hopper, W = Walker, r = random, m = medium, mr = medium-replay, me = medium-expert, AVG.=average.}
 \label{fig: analysis}
\end{figure*}

We find that directly finetuning offline RL policies with online interactions remains inefficient in many cases, as shown in Figure~\ref{fig:finetune_with_online_curve}. 
Most existing works~\citep{adaptiveBC,off2on} assume that the above issue results from the state-action distribution shift between the offline and online samples.
To handle this distribution shift, they propose different methods to progressively transfer from offline to online finetuning settings, including the balance between RL objective and pessimism~\citep{adaptiveBC} and between the usage of offline and online samples~\citep{off2on}.

The main contribution of this paper is to offer a comprehensive understanding of the problem in O2O RL, focusing on \textbf{Q-value estimation}, a perspective that has been underexplored. While prior research acknowledges this issue, our work delves deeper into it, unveiling two key challenges: (1) \textit{biased Q-value estimation} and (2) \textit{inaccurate rank of Q-value estimation}, which represents the distinguishability in the quality of different state-action pairs.
Despite most offline RL policies trained in pessimistic manners such as conservative Q-learning~\citep{cql}, action constraints~\citep{td3_bc}, and Q-ensemble~\citep{edac}, we still observe severe overestimation (CQL and TD3-BC) or underestimation (EDAC) compared to an online RL policy trained from scratch with similar performance. These disparities result in in a severe bootstrap error, as shown in Figure~~\ref{fig:percent_diff}.
Moreover, it is well-known that the inaccurate rank of Q-value estimation, such as inaccurate advantage, results in misleading signals to policy updates. 
To quantitatively assess the accuracy of Q-value estimation rank, we compute Kendall’s $\tau$ coefficient between the estimated and true Q-values, separately for offline RL and online RL policies. Figure~\ref{fig:kendall} shows that offline RL methods (CQL, TD3-BC, and EDAC) have lower Kendall’s $\tau$ coefficient compared to the online RL, SAC. This indicates that the offline RL is less accurate in ranking estimated Q-values than the online RL. Consequently, the signal from the inaccurate Q-value estimation used in policy updates is misleading, which causes instability and slow improvement of performance during online fine-tuning.

\begin{figure*}[t]
    \centering
    \includegraphics[width=0.99\linewidth]{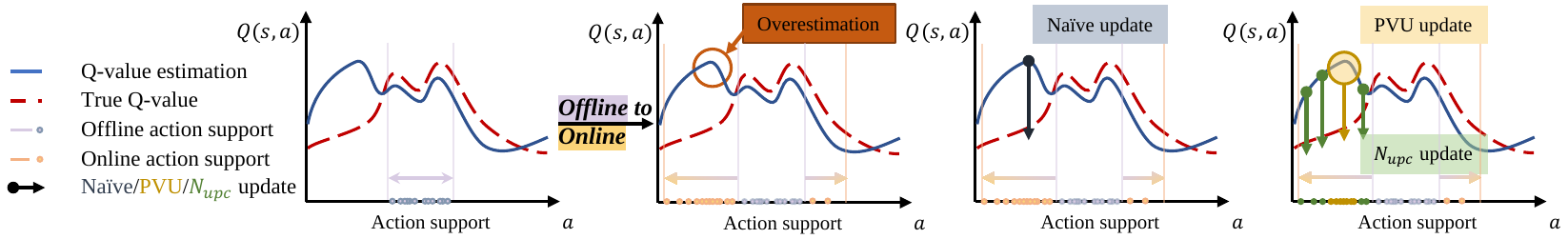}
    \caption{
    Comparison of SO2 against baseline methods. The \textcolor[RGB]{245,213,134}{orange} and \textcolor[RGB]{214,203,225}{purple} spots represent offline and online dataset samples, respectively. The two figures on the left exhibit that from \textit{offline to online} the expanding action support introduces unseen samples with the biased Q-value.
    On the right side, we illustrate the basic idea of SO2 compared with the naive update. 
    Specifically, Perturbed Value Update (PVU) (\textcolor[RGB]{184,146,48}{yellow}) provides indirect \textcolor[RGB]{184,146,48}{range-based} regularization leading to smoothed Q-value estimation.
    The $N_{upc}$ (\textcolor[RGB]{94,129,63}{green}) provides \textcolor[RGB]{94,129,63}{more frequency of Q-value updates} after each online collection. Both of them further improve the accuracy of Q-value estimation during online finetuning.
    }
     \label{fig:mainfigure}
     \vspace{-2ex}
\end{figure*}

To mitigate the Q-value estimation problems stated above, we identify two widely used techniques:  
(1) perturbed value update and (2) increased frequency of Q-value updates.
Specifically, we perturb the target action with an extra noise to smooth out the biased Q-value estimation with sharp peaks. 
This prevents the agent from overfitting to a specific action that might have worked in the past but might not generalize well to new situations. By incorporating noisy actions, the agent is encouraged to explore different actions in the next state, reducing overestimation bias. This encourages the agent to consider a range of plausible actions rather than fixating on a single seemingly optimal action. Consequently, it leads to more accurate value estimates, mitigating value estimation bias in online RL.
Additionally, we remark that the biased Q-value estimation requires an increased frequency of updates to converge rapidly to a normal level.
Increasing the update frequency for Q-values and policies makes the learning process more responsive to new experiences, resulting in more accurate value estimates and faster convergence towards the optimal policy.
As illustrated in Figure~\ref{fig:mainfigure},
both of them provide more accuracy of Q-value function estimation to further improve performance during online finetuning.

The proposed \textbf{S}imple method for \textbf{O}2\textbf{O} RL, named SO2,  is evaluated on (1) MuJoCo locomotion tasks, and (2) dexterous manipulation tasks, using an offline policy pretrained on the D4RL. Our experiments highlight that SO2 consistently outperforms prior state-of-the-art methods in sample efficiency and asymptotic performance on various tasks and datasets.

\section{Background}

Reinforcement learning aims to learn a policy that maximizes the expected cumulative discounted reward in the Markov Decision Process (MDP), which is defined by $(\mathcal{S},\mathcal{A}, \mathcal{R}, p, \gamma)$, with the state space $\mathcal{S}$, the action space $\mathcal{A}$, reward function $\mathcal{R}$, transition dynamics $p$, and discount factor $\gamma \in \left(0, 1 \right)$. 
The actor-critic algorithms maintain the Q-value function $Q\left(\mathbf{s}, \mathbf{a}\right)$ and the learned policy $\pi(\mathbf{a}\mid \mathbf{s})$.

\paragraph{Offline RL.}
Offline RL has an assumption that the agent leverages the logged datasets without any interaction with the environment before the evaluation and deployment. In the Offline RL setting, the agent is trained in a logged dataset $\mathcal{D}$ which is sampled from the environment by a behavior policy $\pi_{\beta}(\mathbf{a}\mid \mathbf{s})$. For the actor-critic algorithm given a fixed dataset $\mathcal{D} = \left\{ (\mathbf{s}, \mathbf{a}, r, \mathbf{s}^{\prime})\right\}$, we can represent value iteration and policy improvement by:
\begin{gather}
{Q}^{k+1} \leftarrow \arg\min_{Q} \mathbb{E}_{\mathbf{s}, \mathbf{a} \sim \mathcal{D}} \left[ \left({\mathcal{B}}^{\pi^{k}} {Q}^{k}(\mathbf{s}, \mathbf{a})  - Q(\mathbf{s}, \mathbf{a}) \right)^2 \right],
\label{eq:bellman}
\\
{\pi}^{k+1} \leftarrow \arg\max_{\pi} \mathbb{E}_{\mathbf{s} \sim \mathcal{D}, \mathbf{a} \sim \pi(\mathbf{a} \mid \mathbf{s})}\left[{Q}^{k+1}(\mathbf{s}, \mathbf{a})\right],
\end{gather}
where the $\mathcal{B}^{\pi}$ is the bellman operator on the fixed dataset following the learned policy  ${\pi} \left(\mathbf{a} \mid \mathbf{s}\right)$, $\mathcal{B}^\pi Q\left(\mathbf{s}, \mathbf{a}\right) = \mathbb{E}_{\mathbf{s}^{\prime} \sim p(\mathbf{s}, \mathbf{a})}\left[ r+\gamma \mathbb{E}_{\mathbf{a}^{\prime} \sim \pi\left(\mathbf{a}^{\prime} \mid \mathbf{s}^{\prime}\right)}\left[\hat{Q}\left(\mathbf{s}^{\prime}, \mathbf{a}^{\prime}\right)\right] \right]$.

\paragraph{Online RL.}
Following a given policy $\pi(\mathbf{a} \mid \mathbf{s})$, the objective of online RL is the expected cumulative discounted reward following the policy in the MDP formulated by $
\mathbb{E}_{\mathbf{a} \sim \pi(\mathbf{a} \mid \mathbf{s})} \sum_{t=0}^{\infty} \gamma^{t}r_{t}$.
Q-Learning methods maintain a Q-function $Q^{\pi}(\mathbf{s}, \mathbf{a})$ to estimate $\mathbb{E}_{\pi}\left[\sum_{t=0}^{\infty} \gamma^{t} r_{t} \mid \mathbf{s}_{0}=\mathbf{s}, \mathbf{a}_{0}=\mathbf{a}\right]$, which measures the expected discounted return following a given policy $\pi(\mathbf{a} \mid \mathbf{s})$.

\paragraph{O2O RL.}
In O2O RL, we consider a setting where deployed agents are allowed to further improve performance by finetuning with online collected samples. 
This setting contains two phases: (1) offline training and (2) online finetuning. At the offline training phase, the logged datasets are utilized to pretrain the parameters of the policy model. At the online finetuning phase, additional online samples are collected to finetune the parameters of thepretrained policy model.

\section{Q-Value Estimation Issues in O2O RL}
\label{sec: challenge}
In this section, we systematically study the issues in O2O RL by focusing on Q-value estimation.  
We begin by assessing the performance of offline RL methods in O2O scenarios and subsequently pinpoint the root cause as Q-value estimation.

\paragraph{Performance.} We evaluate the performance of both standard online RL methods and offline RL methods with online finetuning. 
The online RL methods includes SAC and the online variant of EDAC, which are trained without any offline dataset. 
Their curves are labeled with star marks in Figure~\ref{fig:finetune_with_online_curve}.
The offline RL methods include policy constraints (TD3-BC), conservative Q-value learning (CQL), and Q-ensemble (EDAC) in the O2O RL setting. 
Additionally, we evaluate the loose variant of those methods by removing the pessimistic constraints during online finetuning after offline pretraining. 
The performance of standard online RL methods and offline RL methods is averaged on three environments (HalfCheetah, Hopper, and Walker2D). Particularly, for offline RL, each environment has four datasets with different qualities in D4RL benchmark and generates four pretrained policies, resulting total 12 policies to average on.
It is observed in Figure~\ref{fig:finetune_with_online_curve} that the improvement of CQL, EDAC, TD3-BC and the loose variant of CQL and EDAC is slow in online finetuning; and the asymptotic performance of the loose variant of TD3-BC is significantly worse than online RL and its initial performance.
In summary, while offline algorithms (CQL, TD3-BC and EDAC) all achieve excellent performance in offine settings, all of them achieve slow performance improvement or even performance degradation during online finetuning.

\paragraph{Normalized difference of the Q-value.}
We use normalized difference, also called percent difference in~\citep{td3_bc}, to measure the difference between the estimated Q-value and the true Q-value, formalized by $\frac{Q^{estimated} \ - \ Q^{true}}{Q^{true}}$.
$Q^{true}$ is computed based on actual returns obtained along an adequately extended trajectory collected by the current policy, providing an accurate reflection of the true Q-value.
To highlight the difference between online RL and offline RL algorithms, the normalized differences of offline RL algorithms are subtracted by the normalized difference of the online RL baseline, SAC~\cite{sac}. Therefore, a positive normalized difference means that the Q-value estimation of the offline RL algorithm tends to overestimate, compared to the online baseline, and vice versa.
Figure~\ref{fig:percent_diff} demonstrates the subtracted normalized difference of CQL, TD3-BC and EDAC.
While offline algorithms deal with the out-of-distribution problem via a pessimistic perspective, TD3-BC and CQL usually overestimate the Q-value for the unseen state-action pairs collected online. 
Conversely, EDAC faces the issue of Q-value estimation bias, notably leading to significant Q-value underestimation when compared to online reinforcement learning algorithms. 
This Q-value estimation bias propagates through the Bellman Equation, resulting in suboptimal policy updates or, in some cases, divergence. Specifically, excessively high Q-value estimates can lead to the collapse of Q-value, while overly low Q-value estimates can impede progress, causing slow improvement.

\paragraph{Kendall's $\tau$ coefficient over Q-value.}
Kendall's $\tau$ coefficient measures the rank correlation between two sets of variables. To assess the precision of the rank-ordering of estimated Q-values, we employ Kendall's $\tau$ coefficient to compare them with true Q-values. We evaluate this metric on policies pretrained by CQL, TD3-BC, and EDAC, as well as online SAC, all of which exhibit comparable performance. This evaluation follows a structured procedure: (1) Roll out with each pretrained policy to collect multiple episodes of state-action pairs; (2) Select collections of pairs within each episode using a sliding window approach, denoting each collection as $P_i$, where $i \in [1, ..., M]$ and $M$ represents the window number (10, as used in reported results); 
(3) Compute both the estimated Q-value and true Q-value of all state-action pairs within each $P_i$; 
(4) Compute Kendall's ratio $K_i$ for each collection, and derive the final metric $K$ as the average, given by $K = \frac{1}{M}\sum{K_i}$.
Figure~\ref{fig:kendall} exhibits the $K$ for each RL algorithm. 
The results reveal that offline RL algorithms, CQL, TD3-BC, and EDAC, exhibit significantly lower $K$ values compared to the online RL algorithm SAC.  This suggests that the estimated Q-values are not accurate in assessing the relative qualities of different state-action pairs.

\begin{figure*}[t]
    \centering
    \includegraphics[width=0.99\linewidth]{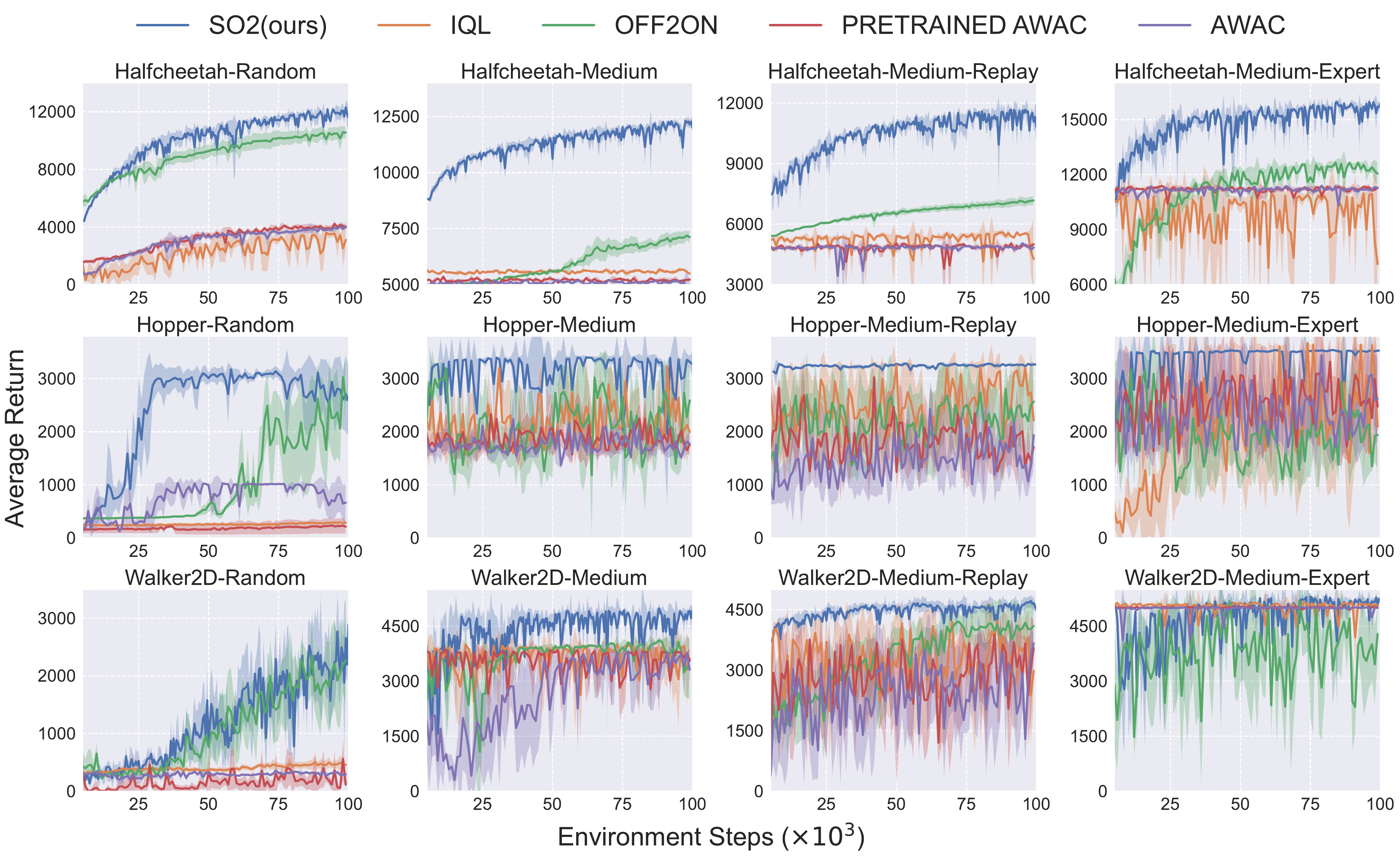}
    \caption{Learning curves comparing the performance of ours against O2O RL baselines pretrained from D4RL datasets with 100k environment steps.Averaged over 4 seeds with the shaded area showing standard deviation. Our proposed method outperforms all baselines by a large margin in terms of both sample efficiency and asymptotic performance with low variance.}
    \label{fig: leanring_curve}
\end{figure*}

\paragraph{Summary.}
While these offline algorithms (CQL, TD3-BC, and EDAC) achieve outstanding performance after pretraining and provide great initial actors for online finetuning, all of them improve slowly with high variances in different environments with different settings, due to the biased estimation and inaccurate rank of Q-value. Thus, how to obtain an accurate Q-value estimation is the bottleneck of O2O RL.

\section{A Simple but Effective Approach to O2O RL}

\subsection{Algorithmic Details}

To provide proper guidance for online policy updates, we present an O2O RL approach that improves the accuracy of Q-value estimation.
Our algorithm builds on top of the SAC with Q-ensemble,
with two modifications, Perturbed Value Update, and Increased Frequency of Q-value Update.

\textbf{Perturbed Value Update (PVU)}. We modify the update step of the ensemble Q-value, as follows:
\begin{equation}
\label{eq: smooth}
\begin{aligned}
\begin{array}{c}
\mathcal{T} Q{\phi_i}(\mathbf{s}, \mathbf{a})
\leftarrow
r+\gamma\left(
 \hat{Q}_{\phi_i}{\left(\mathbf{s^{\prime}}, \mathbf{a^{\prime}} + \epsilon \right)}
-\beta \log \pi\left(\mathbf{a^{\prime}} \mid \mathbf{s^{\prime}}\right)
\right),
\\
\quad \mathbf{a^{\prime}} \sim \pi\left(\cdot \mid \mathbf{s^{\prime}}\right), 
\\
\epsilon 
\sim 
\operatorname{clip}(\mathcal{N}(0, \sigma),-c, c), \ i=1,\ldots,N_{ensemble}, 
\end{array}
\end{aligned}
\end{equation}
where $\epsilon$ represents action noise with variance $\sigma$ and bounded by $c$. $\phi_i$ is the parameter of the $i^{th}$ Q network in the ensemble method, with a total of $N_{ensemble}$ networks.
This perturbation enlarges the action distribution used for estimating the target Q-value, resulting in a smoother Q-value estimate.
Such smooth estimation prevents extremely exploiting the state on which the action has biased Q-value estimation, often seen as sharp peaks.
Hence, this perturbation encourages policy exploration of state-action pairs, even when they have low estimated Q-values inherited from the offline RL pretraining, and further updates the Q-value estimation of these pairs to find potential strategies with high returns.

\textbf{Increased Frequency of Q-value Update.} To alleviate the inaccurate Q-Value estimation in offline RL, as discussed in Sec.~\ref{sec: challenge}, we increase the update frequency of the Q-value function after each online collection. Specifically, we denote the update frequency after each online sample collection as $N_{upc}$, where ${upc}$ means update per collection.

\subsection{Implementation Details}

We describe the practical implementation of SO2 based on SAC with Q-ensemble in DI-engine, a DRL framework utilized by various methods~\cite{ace, MPDP, maskedma, wang2023efficient}.
The pseudocode is shown in Algorithm~\ref{alg: so2}, with differences from Q-ensemble SAC in blue.
After each online collect iteration, we sample a mini-batch $\mathcal{B}$ from the replay buffer. The buffer consists of online samples as well as the abundant samples in the offline dataset, which further improves the stability of the training. Then, we update the Q-value network for $N_{upc}$ times. 
Elaborate details of our algorithm and the baselines are provided in Appendix.

\begin{algorithm}[t!]
    \caption{SO2}
	\label{alg: so2}
	\begin{algorithmic}[1]
	\State Initialized policy parameters $\theta$, Q-function parameters \blue{$\{\phi_{j}\}_{j=1}^N$}, target Q-function parameters $\{\phi_{j}'\}_{j=1}^N$, offline replay buffer $\mathcal{D^{\text{off}}}$, online replay buffer $\mathcal{D^{\text{on}}}$, epoch number $E$ and frequency of Q-value updates $N_{upc}$, and batch size $B$.
\For{epoch $e$ in $1,2,..., E $}
	\State Collect a transition $\tau = (s, a, r, s^{\prime}) $  via environment interaction with $\pi_{\theta}$. \Comment{Collect samples}
	\State Update online replay buffer $\mathcal{D^{\text{on}}} \leftarrow \mathcal{D^{\text{on}}} \cup {\tau}$.
	\For{\blue{\text{$N_{upc}$} iterations}}
	\State Sample a mini-batch $\mathbf{b}$ of transition $\left(\mathbf{s}, \mathbf{a}, r, \mathbf{s'} \right)$
	from \blue{$\mathcal{D^{\text{off}}} \cup \mathcal{D^{\text{on}}}$}.
	\State Compute target Q-values $\blue{\mathcal{T} Q_{\phi_i} \left(\mathbf{s}, \mathbf{a}\right)}$ with the objective in Equation~\ref{eq: smooth}.
	\State Update each Q-function $Q_{\phi_i}$ by gradient descent with:
	\Comment{Value Phase}
    \[
	\nabla_{\phi_i} \frac{1}{B} \sum_{
	\left(
	\mathbf{s}, \mathbf{a}, r, \mathbf{s'}\right) \in \mathbf{b}} \bigg(Q_{\phi_{i}}\left(\mathbf{s}, \mathbf{a}\right)
	-\blue{\mathcal{T} Q_{\phi_i} \left(\mathbf{s}, \mathbf{a}\right)}\bigg)^{2}.
	\]

	\State Update policy by gradient ascent with: \Comment{Policy Phase}

 \begin{adjustbox}{max width=0.85\linewidth}
	\begin{varwidth}{\linewidth-\algorithmicindent}
	\[
	\nabla_{\theta} \frac{1}{B} \sum_{\mathbf{s} \in \mathbf{b}}\left({\min _{j=1,\ldots,N} Q_{\phi_{j}}\left(\mathbf{s}, \tilde{\mathbf{a}}_\theta (\mathbf{s})\right)}-\beta \log \pi_{\theta}\left(\tilde{\mathbf{a}}_\theta (\mathbf{s}) \mid \mathbf{s}\right)\right),
	\]
	where $\tilde{\mathbf{a}}_\theta(\mathbf{s})$ is a sample from $\pi_\theta(\cdot \mid \mathbf{s})$ which is differentiable w.r.t.\ $\theta$ via the reparametrization trick.
	\end{varwidth}
 \end{adjustbox}
	\vspace{0.1em}
	\State Update target networks with $\phi_i' \leftarrow \rho \phi_i'+(1-\rho) \phi_i$.
\EndFor
\EndFor
	\end{algorithmic}
\end{algorithm}

\section{Experiments}
\label{sec:exp}

Our experimental evaluation is aimed at studying the following research problems:
\begin{itemize}[leftmargin=5.5mm]
    \item [$\bullet$] \textbf{Performance:} Can our method improve the performance compared to existing O2O RL approaches and online RL approaches trained from scratch (see Figure~\ref{fig: leanring_curve})?
    \item [$\bullet$] \bm{$N_{upc}:$} Does increasing update frequency per collection effectively enhance performance (see Table~\ref{tab: upc})?
    \item [$\bullet$] \textbf{PVU:} Does the proposed Perturbed Value Update stabilize the O2O training (see Table~\ref{tab: eta})? 
    \item [$\bullet$] \textbf{Q-value estimation:} Whether the proposed method can effectively address Q-value estimation issues including estimation bias and inaccurate rank (see Figure~\ref{fig:kendall_ablation})?
    \item [$\bullet$] \textbf{Extension:} Does our method generalize to more challenging robotic manipulation tasks (see Table~\ref{tab: adroit})?
    \item [$\bullet$] \textbf{Compatibility:} Is our method compatible with other O2O RL algorithms (see Appendix)?

\end{itemize}

\begin{table*}[t!]
\vspace{1ex}
\centering
\resizebox{0.99\linewidth}{!}
{
\centering
\begin{tabular}{l|r|r|r|r|r|r|r}
\toprule
\multicolumn{2}{l|}{\textbf{Dataset}} & \multicolumn{1}{c|}{AWAC}		& \multicolumn{1}{c|}{ODT} 		& \multicolumn{1}{c|}{IQL}		& \multicolumn{1}{c|}{Off2ON}		& \multicolumn{1}{c|}{PEX} & \multicolumn{1}{c}{SO2 (Ours)} \\

\midrule
\multirow{3}{*}{\rotatebox{90}{Random}} 
& HalfCheetah & 6.5 $\rightarrow$35.0  & 10.1 $\rightarrow$ 18.3  &13.4 $\rightarrow$ 27.3  &27.7 $\rightarrow$ 87.2  &15.6 $\rightarrow$ 55.3  &37.7 $\rightarrow$ \textbf{95.6}  \\
& Hopper & 6.7 $\rightarrow$21.1  & 6.9  $\rightarrow$ 31.2   &7.6  $\rightarrow$ 9.3   &10.5 $\rightarrow$ \textbf{80.4}  &11.0  $\rightarrow$ 47.0 &9.2  $\rightarrow$ \textbf{79.9} \\
& Walker2d & 5.9 $\rightarrow$6.3   & 6.4  $\rightarrow$ 12.3   &6.8  $\rightarrow$ 9.9   &10.3 $\rightarrow$ 49.4  &8.9  $\rightarrow$ 15.4 &6.9  $\rightarrow$ \textbf{62.9}  \\
\midrule

\multirow{3}{*}{\rotatebox{90}{\begin{tabular}[c]{@{}c@{}}Medium\\ Replay\end{tabular}}}
& HalfCheetah & 40.5  $\rightarrow$ 41.2  	& 32.4  $\rightarrow$ 39.7 	& 42.7  $\rightarrow$ 36.7  	& 42.1  $\rightarrow$ 60.0  	& 45.5  $\rightarrow$ 51.3 & 62.5  $\rightarrow$ \textbf{89.4} \\
& Hopper 	    & 37.7  $\rightarrow$ 60.1  	& 60.4  $\rightarrow$ 78.5  	& 75.8  $\rightarrow$ 68.5  	& 28.2  $\rightarrow$ 79.5  	& 31.5  $\rightarrow$ 97.1 & 97.0  $\rightarrow$ \textbf{101.0} \\
& Walker2d 	& 24.5  $\rightarrow$ 79.8  	& 44.2  $\rightarrow$ 71.8  	& 75.6  $\rightarrow$ 64.9  	& 17.7  $\rightarrow$ 89.2  	& 80.1  $\rightarrow$ 92.3 & 80.9  $\rightarrow$ \textbf{98.2} \\

\midrule

\multirow{3}{*}{\rotatebox{90}{Medium}}
& HalfCheetah & 43.0  $\rightarrow$ 42.4  	& 42.7  $\rightarrow$ 42.1  	& 46.7  $\rightarrow$ 46.5  	& 39.3  $\rightarrow$ 59.6  	& 50.8  $\rightarrow$ 60.9 & 73.3  $\rightarrow$ \textbf{98.9} \\
& Hopper 	    & 57.8  $\rightarrow$ 55.1  	& 47.2  $\rightarrow$ 67.0  	& 73.4  $\rightarrow$ 61.9  	& 97.5  $\rightarrow$ 80.2  	& 56.5  $\rightarrow$ 87.5 & 77.6  $\rightarrow$ \textbf{101.2} \\
& Walker2d 	& 35.9  $\rightarrow$ 72.1  	& 72.0  $\rightarrow$ 72.2  	& 84.7  $\rightarrow$ 78.8  	& 66.2  $\rightarrow$ 72.4  	& 80.1  $\rightarrow$ 92.3 & 76.4  $\rightarrow$ \textbf{107.6}\\

\midrule
\multirow{3}{*}{\rotatebox{90}{\begin{tabular}[c]{@{}c@{}}Medium\\ Expert\end{tabular}}}
& HalfCheetah & 65.2  $\rightarrow$ 93.0  	& 47.1  $\rightarrow$ 94.7  	& 88.2  $\rightarrow$ 59.6  	& 56.7  $\rightarrow$ 99.3  	& 37.3  $\rightarrow$ 76.2 & 87.1  $\rightarrow$ \textbf{130.2} \\
& Hopper 	    & 55.5  $\rightarrow$ 81.3  	& 64.9  $\rightarrow$ 111.7  	& 52.8  $\rightarrow$ 65.0  	& 112.0  $\rightarrow$ 60.2  	& 91.3  $\rightarrow$ 99.3 & 67.2  $\rightarrow$ \textbf{109.1} \\
& Walker2d 	& 108.3  $\rightarrow$ 108.7  	& 78.9  $\rightarrow$ 107.8  	& 110.8  $\rightarrow$ 110.4  	& 102.1  $\rightarrow$ 93.3  	& 83.3  $\rightarrow$ \textbf{114.2}  & 109.5  $\rightarrow$ \textbf{112.9} \\

\midrule
\multicolumn{1}{l}{} 
&\textbf{Average} & 40.6  $\rightarrow$ 58.0	 
                & 42.8	 $\rightarrow$ 62.3
                & 56.6	 $\rightarrow$ 53.2     
                & 48.4 $\rightarrow$  75.9
                & 46.0  $\rightarrow$ 75.6  
                & 76.1  $\rightarrow$ \textbf{98.9} \\
\bottomrule
\end{tabular}
}
\caption{Normalized average returns on D4RL Gym tasks, averaged over 4 random seeds. 
SO2 significantly improves the performance compared to the state-of-the-art OFF2ON, despite being much simpler to implement.}
\end{table*}

\begin{figure*}[t!]
    \centering
    \includegraphics[width=0.99\linewidth]{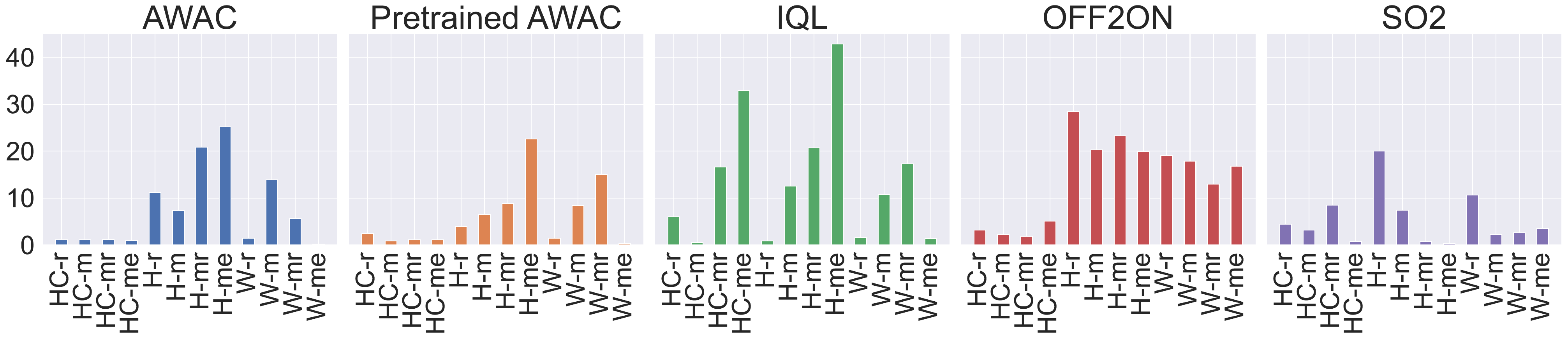}
    \caption{Comparing the standard deviation of ours against of O2O RL baselines pretrained from D4RL datasets with 100k environment steps. The standard deviation is averaged over 4 seeds. Our proposed method SO2 has a smaller standard deviation compared with existing work.}
    \label{fig: variance}
\end{figure*}

\subsection{Evaluation on MuJoCo tasks}

\paragraph{Setup.} We first evaluate SO2 and baselines O2O RL algorithms on MuJoCo~\cite{mujoco} tasks trained from the D4RL-v2 dataset consisting of three environments including HalfCheetah, Walker2d and Hopper, each with four level datasets collected by policies with different levels, including Random, Medium, Medium-Replay and Medium-Expert. We report the performance on the standard normalized return metric in D4RL, averaged over 4 seeds.

\paragraph{Comparative evaluation.}
We take the following methods as our baselines for comparison:
\begin{itemize}
    \item [$\bullet$] \textbf{OFF2ON: } a pessimistic Q-ensemble RL method that proposes a balanced replay to encourage the use of near-on-policy samples from the offline dataset. We use the official implementations for OFF2ON.
    \item [$\bullet$] \textbf{IQL-ft: } an offline method with strong finetuning performance by approximating the upper bound of the Q-value distribution and extracting the policy from the Q-value estimation. We use the implementation in rlkit for IQL.
    \item [$\bullet$] \textbf{AWAC:} an O2O RL method that trains the policy to imitate actions with high advantage estimates. We use the official implementations for it.
    \item [$\bullet$] \textbf{ODT~\cite{odt}:} an O2O RL algorithm based on sequence modeling that considers offline and online finetuning in a unified framework.
     \item [$\bullet$] \textbf{PEX~\cite{pex}:} an RL algorithm based on a policy expansion where the policy set includes offline policy and another policy used for further learning.
\end{itemize}

We train policies for 100K environment steps and evaluate every 1000 environment steps.
Our method uses perturbation noise with $\sigma = 0.3$, $c = 0.6$, and ${N_{upc}} = 10$ as the default setup. 
The complete details are reported in the Appendix.

\begin{figure*}[t!]
  \centering
 
\subfloat[Learning curve]{
\begin{minipage}{0.42\textwidth}
    \includegraphics[width=1.0\linewidth]{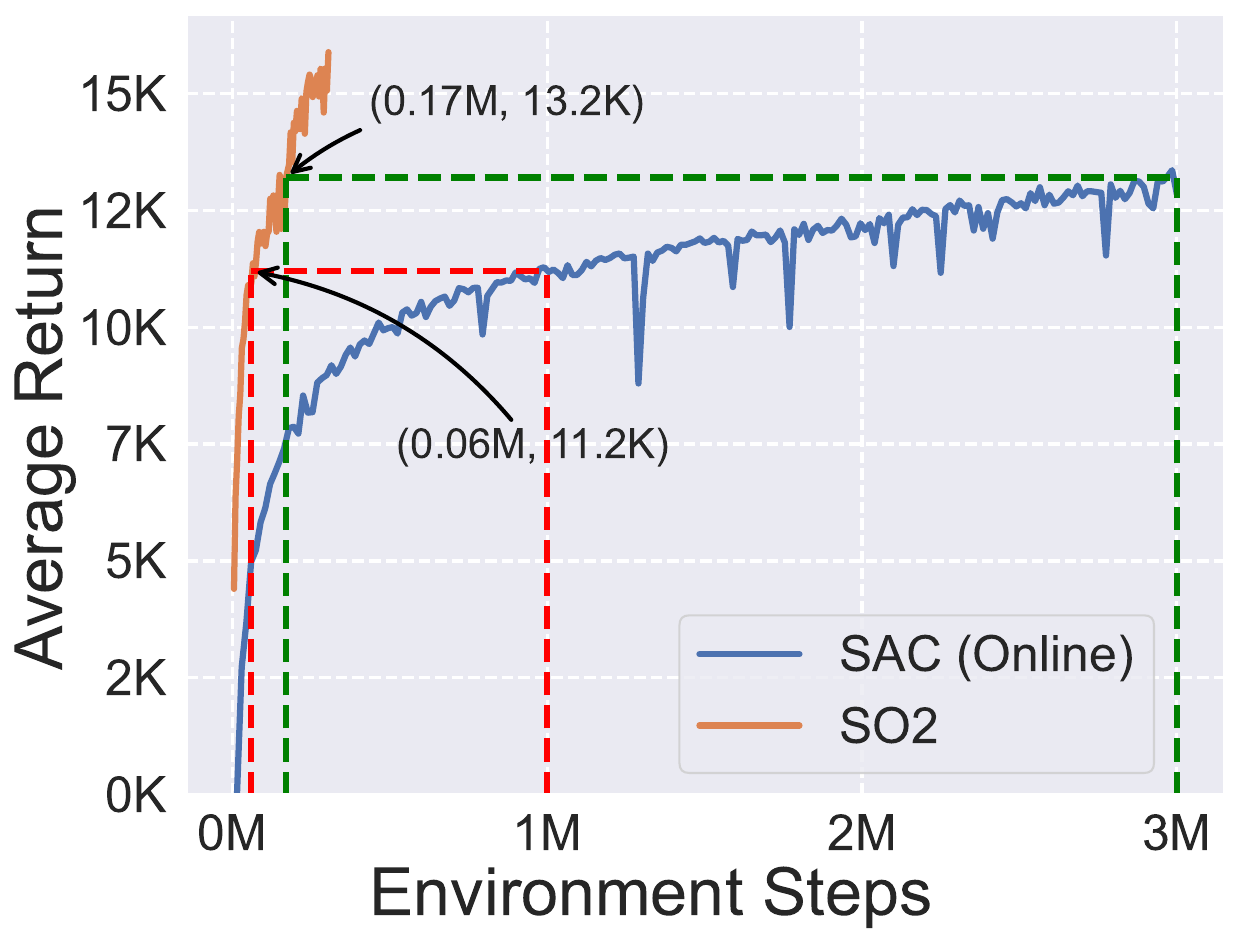}
    \label{fig:halfcheetah-random}
\end{minipage}
\vspace{-3ex}
}
\subfloat[Kendall for Q-value]{
\begin{minipage}{0.42\textwidth}
    \includegraphics[width=1.0\linewidth]{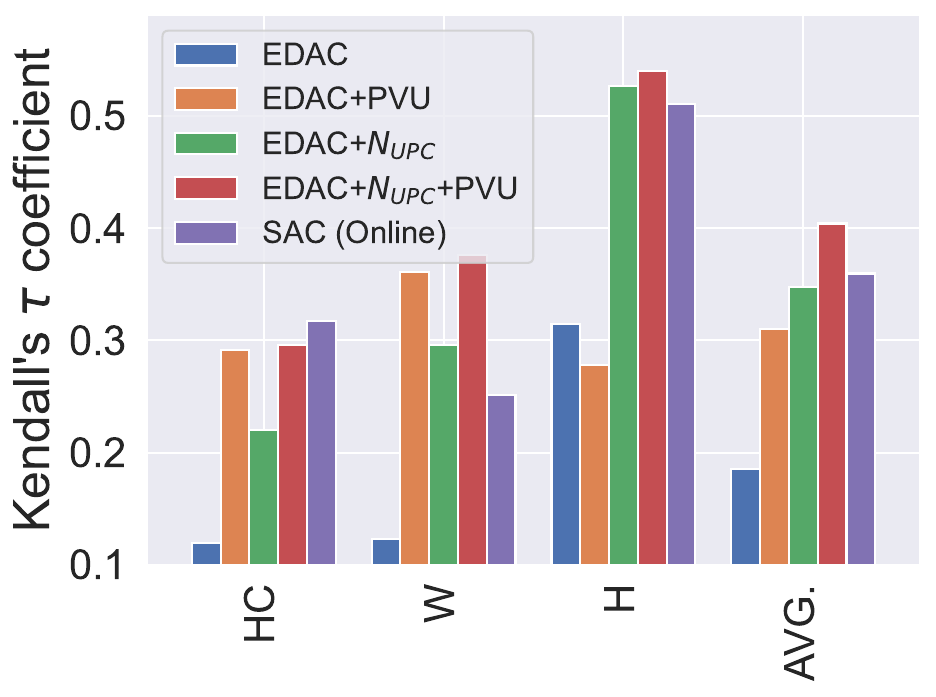}
    \label{fig:kendall_ablation}
\end{minipage}
\vspace{-2.3ex}
}
  \caption{\textbf{(a):} Performance comparing our proposed method against online SAC in HalfCheetah. Although the baseline is initialized from the HalfCheetah-Random dataset with limited performance, our proposed method boosts the baseline algorithm effectively and significantly improves the sample efficiency by up to 17 compared to SAC.
  \textbf{(b)} Kendall's $\tau$ coefficient comparing ablation over Perturbed Value Update (PVU), and update per collection ($N_{upc}=10 $) in the MuJoco environment.
  This measures the rank correlation between true Q-value and Q-value estimation for similar state-action pairs.
HC = HalfCheetah, Hop = Hopper, W = Walker, r = random, m = medium, mr = medium-replay, me = medium-expert.}
 \label{fig: kendall_and_halfcheetah}
\end{figure*}

Figure~\ref{fig: leanring_curve} shows the performance of our proposed SO2 compared with the state-of-the-art O2O algorithms during the online finetuning phase. We re-run the baseline algorithms to make sure a fair comparison.
It is observed that SO2 has the three advantages: 
(1) It outperforms all baselines by a large margin in terms of both sample efficiency and asymptotic performance; 
(2) Compared with baselines, SO2 typically has small variances per environment and per offline dataset, as shown in Figure~\ref{fig: variance};
(3) Even when the offline training data is collected by random policy, SO2 still achieves expert performance with only a few online interactions. 
Specifically, we compare SO2 and SAC, a standard off-policy algorithm, as shown in Figure~\ref{fig:halfcheetah-random}.
Although the baseline is pretrained on the 1M offline data in the HalfCheetah-Random dataset, which has quite limited performance, SO2 only requires 0.17M online environment steps to achieve outstanding performance with 13000 average episodic return, totally 1.17M. In comparison, the standard SAC requires more than 3M environment steps to achieve the same performance. It means we fully release the potential of O2O RL in sample efficiency, even when the most data is randomly collected and with poor quality. This is not observed by any baseline algorithms.

Figure~\ref{fig: leanring_curve} shows the differences in starting performance across different methods arise from the fact that each offline-to-online algorithm utilizes distinct baselines. Moreover, the hyperparameters employed in offline-to-online algorithms are meticulously tailored to their respective baseline results, rendering them inherently sensitive. Achieving consistent starting performance across all algorithms would necessitate an extensive search for suitable parameters, consuming significant computational resources. Consequently, most offline-to-online algorithms~\cite{cql, awac, off2on, odt, pex} report results using different starting points to accommodate these algorithm-specific sensitivities.

\begin{table*}[t!]
    \centering
    \begin{tabular}{l|r|r|r|r}
        \toprule
        Dataset & $\sigma=0.0$ & $\sigma=0.15$ & $\sigma=0.3$ & $\sigma=0.45$ \\
        \midrule
Walker2d-Random          &24.0 	$\pm$15.9  &24.6 	$\pm$12.9 	&\textbf{62.9 	$\pm$10.6} 	&22.9 $\pm$	12.8 \\
Walker2d-Medium          &100.0 $\pm$16.1  &93.5 	$\pm$16.0 	&\textbf{107.6 	$\pm$2.3} 	&105.5$\pm$ 	3.3 \\
Walker2d-Medium-Replay   &73.7 	$\pm$32.3  &94.5 	$\pm$7.7 	&\textbf{98.2 	$\pm$2.6} 	&91.4 $\pm$	3.1 \\
Walker2d-Medium-Expert   &108.2 $\pm$12.1  &110.6 	$\pm$6.9 	&\textbf{112.9} 	$\pm$3.5 	&110.2$\pm$ 	\textbf{1.9} \\
\midrule
Average                  &76.5 	$\pm$19.1  &80.8 	$\pm$10.9 	&\textbf{95.4 	$\pm$4.7} 	&82.5 $\pm$	5.3 \\
        \bottomrule
    \end{tabular}
\caption{Performance of SO2 over various $\sigma$ pretrained on the D4RL Walker2d datasets.}
\label{tab: eta}
\end{table*}

\paragraph{Analysis on Perturbed Value Update.}
In this paper, we argue that in the O2O task, adding appropriate noise to the target action can make training more stable and results in better policy performance. 
To verify the effectiveness of target noise, we conduct an ablation over $\sigma$.
As shown in Table~\ref{tab: eta}, the baseline without the target noise ($\sigma=0$) has significant variances in all offline datasets, up to 1/3 of the average performance. 
The high variance reveals that even if the learned policies have the same initial performance, some of them may still perform poorly on some episodes, and the learned policy may fluctuate dramatically among each evaluation.
On the contrary, the volatile policy, once equipped with the Perturbed Value Update, achieves consistently impressive performance with low variance across various PVUs.

\begin{table*}[t!]
    \centering
    \begin{tabular}{l|r|r|r}
        \toprule
        Dataset & ${N_{upc}}=1$ & ${N_{upc}}=5$ & ${N_{upc}}=10$\\
        \midrule
        Walker2d-Random          &29.5    $\pm$12.6	&30.8	$\pm$	14.9	&\textbf{62.8	$\pm$	10.6}		   \\
Walker2d-Medium          &97.5    $\pm$	5.0	&97.1	$\pm$	17.4	&\textbf{107.6	$\pm$	2.2}	    \\
Walker2d-Medium-Replay   &94.4    $\pm$	2.2	&\textbf{99.0	$\pm$	1.3}	    &\textbf{98.2	}	$\pm$	2.5    \\
Walker2d-Medium-Expert  &103.7    $\pm$	3.4	&111.2	$\pm$	\textbf{1.4}	    &\textbf{112.9}	$\pm$	3.5	    \\
\midrule
Average                  &81.3 	  $\pm$ 5.8	&84.5	$\pm$	8.7	    &\textbf{95.4}	$\pm$	4.7  \\
        \bottomrule
    \end{tabular}
\caption{Performance of SO2 over various \textbf{$N_{upc}$} pretrained on the D4RL Walker2d datasets.}
\label{tab: upc}
\end{table*}

\paragraph{Analysis on $N_{upc}$.}
To verify the effectiveness of $N_{upc}$, we report the performance of our algorithm on Walker2D with different $N_{upc}$. Table~\ref{tab: upc} shows that the average performance consistently improves with the increase of $N_{upc}$, in all the four sub-environments on Walker2D steadily. 
It is worth noting that increasing $N_{upc}$ on random data has the most obvious effect.
Intuitively, the $N_{upc}$ term induces the value network to favor frequent updates after each interaction with environments. Therefore, the increased $N_{upc}$ leads to a more accurate Q-value estimation, which is more desirable if the initial estimation is sub-optimal.

\paragraph{Analysis on Q-value estimation.}
Figure~\ref{fig:kendall_ablation} shows an accuracy comparison of rank of the estimated Q-value during online finetuning, between the original EDAC and the variants incorporating each components of SO2.
It is shown that our approach outperforms the baseline consistently in all environments, which further validates our idea of target noise and frequent updates to boost Q-value estimation to advance the online finetuning.

\subsection{Evaluation on Adroit tasks}
\paragraph{Setup.}
We also conduct experiments on Adroit, which requires controlling a 24-DoF robotic hand to perform tasks such as Pen and Door. Specifically, we report the performance of offline pretraining and online finetuning, measured by normalized average scores. The offline policy is trained with limited human demonstrations in the D4RL-v0 dataset, and the online finetuning policy is trained with \textbf{1M} environment steps for AWAC, CQL, and IQL, and with only \textbf{400k} environment steps for ours.
Results for AWAC, CQL and IQL are quoted from IQL paper~\cite{iql}. The complete experimental details are provided in Appendix.

\vspace{-1.2mm}
\paragraph{Comparative evaluation.}
Table~\ref{tab: adroit} shows that our method also outperforms all other algorithms by a large margin. 

\begin{table*}[t!]
    \centering
    
    \begin{tabular}{l|l|l|l|l}
        \toprule
        Dataset       & AWAC                               & CQL & IQL & Ours(400k) \\ 
        \midrule
        pen-human-v0  & 44.6$\rightarrow$70.3 &31.2$\rightarrow$9.9     &37.4$\rightarrow$60.7     & 51.4$\rightarrow$\textbf{70.6}  \\
        door-human-v0 & 1.3$\rightarrow$30.1  &0.2$\rightarrow$0.0     &0.7$\rightarrow$32.3     & 16.5$\rightarrow$\textbf{66.2}  \\  
        \midrule
        Total         
        &46.7$\rightarrow$103.1 &31.5$\rightarrow$9.9 &38.1$\rightarrow$124.0 &67.9$\rightarrow$ \textbf{136.8} \\
        \bottomrule
    \end{tabular}
\caption{Normalized average scores on the D4RL Adroit tasks, including pretrained from the offline dataset and after online finetuning. We report the online finetuning results of AWAC, CQL, and IQL with \textbf{1M} environment steps, while the online finetuning results of ours with \textbf{400k} steps.
}
\label{tab: adroit}
\end{table*}

\section{Related Work}

\paragraph{Offline RL}
Offline RL algorithms employ static datasets to train agents but face Q-value overestimation challenges, especially with unseen state-action pairs~\cite{bcq}.
Solutions can be categorized as follows:
(1) Policy constraint methods~\cite{awac, td3_bc, wu2019behavior} enforce policy alignment with the behavior policy in logged datasets, minimizing policy distribution shift through behavior cloning and KL-divergence; 
(2) Pessimistic value methods~\cite{cql} regularize the Q-value function to mitigate the overestimation issues in the OOD state-action pairs by adding a conservative penalty to the RL object;
(3) Uncertainty-based methods~\cite{edac, BEAR, pbrl} measure the distribution shift using Q-value uncertainty estimations to provide a robust signal for policy updates.

\paragraph{Online RL with ofﬂine datasets.}
Several works~\cite{ijspeert2002learning, kim2013learning, zhu2019dexterous, nair2017combining, theodorou2010generalized, gupta2020relay} have explored improving online RL by utilizing offline datasets, assuming these datasets contain optimal demonstrations specific to the current environment.
However, real-world offline datasets often come from various sources and contain large amounts of sub-optimal demonstrations, such as autonomous driving and manipulation. These methods may not perform well in such scenarios due to estimation bias from sub-optimal datasets.
An alternative approach~\cite{awac, off2on} takes a pessimistic view, avoiding the assumption of optimality. 
These methods focus on the challenge of significant distribution gaps between offline datasets and online samples during offline-to-online finetuning.
Particularly, AWAC~\cite{awac} enforces an implicit constraint on policy updates to align the learned policy with policies in both offline and online samples.
~\cite{off2on} employs Q-ensemble and balanced replay, encouraging the use of near-on-policy samples from offline demonstrations based on online-ness predictions by a neural network.
AdaptiveBC~\cite{adaptiveBC} proposes randomized
ensemble of Q-function and dynamically balances the RL objective with behavior cloning based on agent performance and training stability.
IQL~\cite{iql} achieves strong performance in finetuning tasks by approximating the upper bound of Q-values for each state and extracting the policy from the Q-value estimation.
ODT~\cite{odt} proposes an O2O RL algorithm based on sequence modeling and entropy regularizers to unify offline pertaining and online finetuning.
PEX~\cite{pex} introduces a policy expansion scheme that includes offline policies along with others for further learning.
QDagger~\cite{qdagger} focuses on optimizing knowledge transfer efficiency within the reincarnating RL paradigm.
While some prior works do acknowledge the problem of inaccurate Q-value estimation, they may not have explicitly dissected the various types of inaccuracies and their specific effects on policy updates and online finetuning performance.
Our research goes beyond existing literature by providing a more comprehensive understanding of these issues within the O2O RL context.
Specifically, we focus on uncovering the challenges related to biased Q-value estimation and inaccurate Q-value ranking inherited from offline RL. 
These issues result in unreliable signals that exacerbate bootstrap errors during finetuning, leading to instability and suboptimal policy updates.
Our investigation and proposed solutions address problems not explored in-depth in previous research.

\section{Conclusion}

In this paper, we have delved into O2O reinforcement learning and systematically studied why this setting is challenging.
Different from most existing works, we in-depth analyze the Q-value estimation issues in offline-to-online including the biased estimation and inaccurate rank of the Q-value, besides the bootstrap error resulting from state-action distribution shift.
Based on this argument, we propose smoothed offline-to-online (SO2). It effectively and efficiently improves the Q-value estimation by perturbing the target action and improving the frequency of Q-value updates.
The proposed method, without any explicit estimation of the state-action distribution shift and complex components to balance offline and online replay buffers, remarkably improves the performance of the state-of-the-art methods by up to 83.1\% on the MuJoco and Adroit environments.

\section{Acknowledgments}
This research is funded by Shanghai AI Laboratory. This work is partially supported by the National Key R\&D Program of China (NO.2022ZD0160100), (NO.2022ZD0160101). This work was done during Yinmin's internship at Shanghai Artificial Intelligence Laboratory. We would like to thank many colleagues for useful discussions, suggestions, feedback, and advice, including: Zilin Wang and Ruoyu Gao.

\newpage

\bibliography{aaai24}

\newpage

\clearpage

\setcounter{section}{0}
\renewcommand\thesection{\Alph{section}} 

{\centering\section*{\LARGE\bf Supplementary Material}}

In this Supplementary Material, we provide more elaboration on the implementation details, results, and analysis.

\section{Additional Implementation Details}
\label{sec:appendix_details1}

\subsection{Environment.} 
We use the following software versions:
\begin{itemize}
    \item CentOS 7.9
    \item Python 3.6
    \item Pytorch 1.5.0~\cite{pytorch}
    \item Gym 0.20.0~\cite{gym}
    \item Mujoco 2.1.2~\cite{mujoco}
\end{itemize}
We leverage the v2 version of D4RL datasets~\cite{d4rl} to provide the pretrained model for AWAC, IQL and ours.
Additionally, we re-run OFF2ON~\cite{off2on} based on the pretrained model provided by the author. 
We conduct all experiments with a single GeForce GTX 1080 GPU and an Intel Core CPU at 2.50GHz.

\subsection{Hyperparameters}
Our experiments of AWAC\footnote{https://github.com/rail-berkeley/rlkit/tree/master/examples/awac}~\cite{awac}, and OFF2ON\footnote{https://github.com/shlee94/Off2OnRL}~\cite{off2on} are conducted on their official implementation provided by their respective authors. And the experiment of IQL~\cite{iql} is based on the re-implementation version of rlkit\footnote{https://github.com/rail-berkeley/rlkit/tree/master/examples/iql}. Our experiments of online RL (\textit{i.g.}, SAC~\cite{sac} and Q-ensemble SAC) and offline RL (\textit{i.g.}, TD3-BC~\cite{td3_bc} and CQL~\cite{cql}) are conducted on DI-engine\footnote{https://github.com/opendilab/DI-engine}~\cite{ding}, a generalized RL framework.

\paragraph{SAC.} We report the online SAC, an off-policy algorithm, based on the implementation of DI-engine and use its default parameters.

\paragraph{CQL.} We report the online finetuning CQL, including the original variant and the loose variant without the conservative penalty, based on the implementation of DI-engine. Specifically, we first use its default parameters to obtain a pretrained model for each environment and each level dataset in D4RL benchmarks, following the protocol of CQL. Based on the pretrained model, we further implement online finetuning via initializing the replay buffer from the offline dataset and updating the policy and Q-function depending on the corresponding RL objectives. For the original variant of CQL, we keep the default min Q weight parameters ($\alpha = 10$) compared to the offline training; for the loose variant of CQL, we update the Q-value network in the SAC manner by removing the min Q weight constraints in Q-function. We outline the hyperparameters used by the original variant of CQL in Table~\ref{tab: CQL_hyperparameters}.

\begin{table}[h]
\centering
\resizebox{1.0\linewidth}{!}{
\begin{tabular}{llllllll}
\toprule
                                     & Hyperparameter                              & Value                          \\
\midrule
\multirow{8}{*}{SAC Hyperparameters} & Optimizer                                   & Adam \\
                                     & Critic learning rate                        & 3e-4                           \\
                                     & Actor learning rate                         & 3e-5                         \\
                                     & Batch size                             & 256                            \\
                                     & Discount factor                             & 0.99                           \\
                                     & Target update rate                          & 5e-3                           \\
                                     & Target entropy                              &  $- \| \text{Action Dim} \| $               \\
                                     & Entropy in Q target                         & False                         \\
\midrule
\multirow{5}{*}{Architecture}        & Critic hidden dim                           & 256                            \\
                                     & Critic hidden layers                        & 3                             \\
                                     & Critic activation function & ReLU
                                     \\
                                     & Actor hidden dim & 256             
                                     \\
                                     & Actor hidden layers                         & 3                           \\
                                     & Actor activation function                   & ReLU                           \\
\midrule
\multirow{3}{*}{CQL Hyperparameters}
                                    & Lagrange                                    & False                          \\
                                    & Min Q weight~$\rho$ (called $\alpha$ in CQL)                                & 10                             \\
                                    & Pre-training steps                          & 0*                          \\
                                    & Num sampled actions             & 10        \\    
\bottomrule                
\end{tabular}
}
\caption{CQL Hyperparameters. We use the hyperparameters following the CQL~\cite{cql}. * denotes the hyperparameter which is different from the original CQL.}
\label{tab: CQL_hyperparameters}
\end{table}

\paragraph{TD3-BC.} We report the online finetuning TD3-BC, including the original variant and the loose variant without the behavior cloning, based on the implementation of DI-engine. Similar to the experiments of CQL, we first obtain a pretrain model on D4RL benchmarks and finetune the pretrained model depending on the corresponding RL objectives. For the original variant of TD3-BC, we keep the default hyperparameter ($\alpha = 2.5$), which controls the balance between RL objective and behavior cloning in TD3-BC; for the loose variant of TD3-BC, we update the Q-value network in the TD3 manner by removing the policy constraints in policy update step. We outline the hyperparameters used by the original variant of TD3-BC in Table~\ref{tab: TD3-BC_hyperparameters}.

\begin{table}[h]
\centering
\resizebox{1.0\linewidth}{!}{
\begin{tabular}{llllllll}
\toprule
& Hyperparameter                              & Value                          \\
\midrule
\multirow{9}{*}{TD3 Hyperparameters} & Optimizer                                   & Adam \\
                                     & Critic learning rate                        & 3e-4                           \\
                                     & Actor learning rate                         & 3e-4                           \\
                                     & Mini-batch size                             & 256                            \\
                                     & Discount factor                             & 0.99                           \\
                                     & Target update rate                          & 5e-3                           \\
                                     & Policy noise                                & 0.2                            \\
                                     & Policy noise clipping                       & (-0.5, 0.5)                    \\
                                     & Policy update frequency                     & 2                              \\
\midrule
\multirow{6}{*}{Architecture}        & Critic hidden dim                           & 256                            \\
                                     & Critic hidden layers                        & 2                              \\
                                     & Critic activation function  & ReLU                      \\
                                     & Actor hidden dim & 256 \\
                                     & Actor hidden layers                         & 2                              \\
                                     & Actor activation function                   & ReLU                           \\
\midrule
TD3-BC Hyperparameters               & $\alpha$                                 & 2.5                            \\
\bottomrule                
\end{tabular}
}
\caption{TD3-BC Hyperparameters. We use the hyperparameters following the TD3-BC~\cite{td3_bc}. * denotes the hyperparameter which is different from the original TD3-BC.}
\label{tab: TD3-BC_hyperparameters}
\end{table}

\paragraph{SO2.} 
We report online finetuning SO2 depending on the Q-ensemble SAC pretrained model. Specifically, depending on the official implementation of EDAC, we first train the Q-ensemble SAC model in the EDAC manner on the D4RL benchmarks, following the protocol of EDAC. Based on the pretrained model, we further implement online finetuning via the SAC-N method where the value update in the EDAC manner without Ensemble Similarity (ES) metric in the DI-engine. We outline the hyperparameters used by SO2 on MuJoco in Table~\ref{tab:edac_gym_params} and on Adroit in Table~\ref{tab:edac_adroit_params}.

\begin{table}[h]
	\centering
	\small
	\begin{adjustbox}{max width=\columnwidth}
		\begin{tabular}{l|c}
			\toprule
			\textbf{Task Name} & \textbf{SAC-$N$} ($N$) \\
			\midrule
			halfcheetah-random & 10  \\
			halfcheetah-medium & 10  \\
			halfcheetah-medium-expert & 10  \\
			halfcheetah-medium-replay & 10  \\
			\midrule
			hopper-random & 50 \\
			hopper-medium & 50 \\
			hopper-medium-expert & 50 \\
			hopper-medium-replay & 50 \\
			\midrule
			walker2d-random & 10 \\
			walker2d-medium & 10\\
			walker2d-medium-expert & 10 \\
			walker2d-medium-replay & 10 \\
			\bottomrule
		\end{tabular}
	\end{adjustbox}
	\caption{Hyperparameters used in the D4RL MuJoCo Gym experiments for Q-ensemble SAC.}
	\label{tab:edac_gym_params}
\end{table}

\begin{table}[h]
	\centering
	\small
	\begin{adjustbox}{max width=\columnwidth}
		\begin{tabular}{l|c}
			\toprule
			\textbf{Task Name} & \textbf{SAC-$N$} ($N$) \\
			\midrule
			pen-human & 20 \\
			door-human & 50 \\
			\bottomrule
		\end{tabular}
	\end{adjustbox}
  \caption{Hyperparameters used in the D4RL Adroit experiments for Q-ensemble SAC.}
	\label{tab:edac_adroit_params}
\end{table}

\section{Additional Results and Analysis}
\label{sec:appendix_results}
\subsection{Comparison Q-value estimation via Kendall's $\tau$ coefficient with fixed policies}
\begin{figure}[h]
    \centering
    \includegraphics[width=0.99\linewidth]{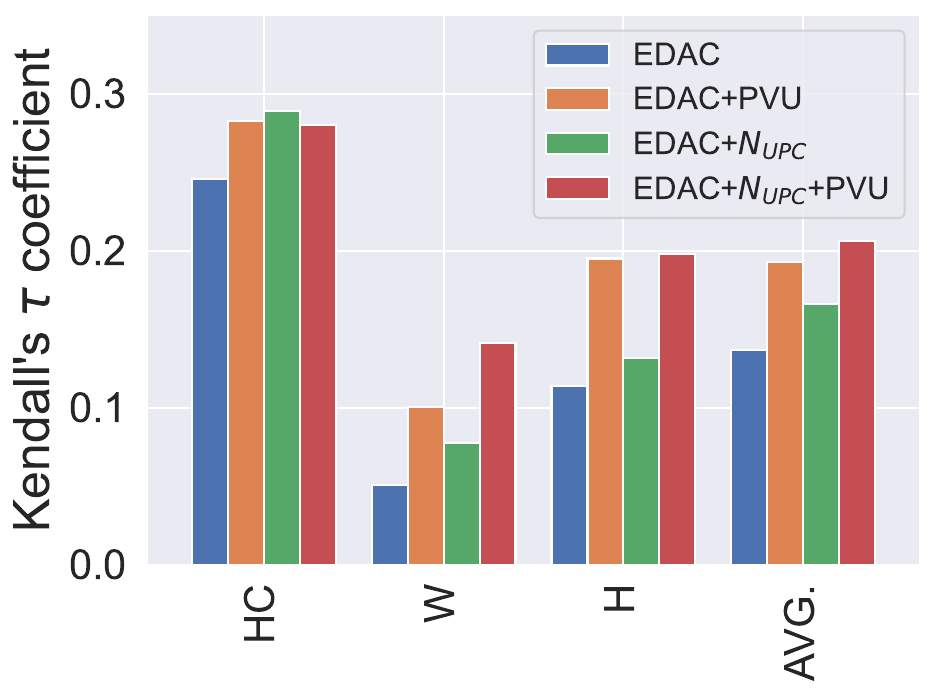}
    \caption{Kendall's $\tau$ coefficient under fixed policies comparing ablation over Perturbed Value Update (PVU), and update per collection ($N_{upc}= 10$) in the MuJoco environment. This measures the rank correlation between true Q-value and Q-value estimation for similar state-action pairs.}
    \label{fig: appendix_kendall_ablation}
\end{figure}

A concern of Kendall's $\tau$ coefficient is that both the different policies and different Q-value estimations have effects on Kendall's $\tau$ coefficient between the true Q-value and Q-value estimation. This makes it unclear whether the high Kendall's $\tau$ coefficient benefits from the improvement of policy or the accurate Q-value estimation.
Therefore, to mitigate the effects of policy updates, we evaluate the Kendall's $\tau$ coefficient after Q-value updates with a similar fixed policy. 
As shown in Figure~\ref{fig: appendix_kendall_ablation}, we find that SO2 (\textit{i.g.}, PVU and $N_{upc}$) generally improves the accuracy of Q-value estimation even with a fixed policy.

\subsection{Comparison SO2 against Online RL and Model-based RL}
\begin{figure*}[h]
\vspace{-2ex}
    \centering
    \includegraphics[width=0.92\linewidth]{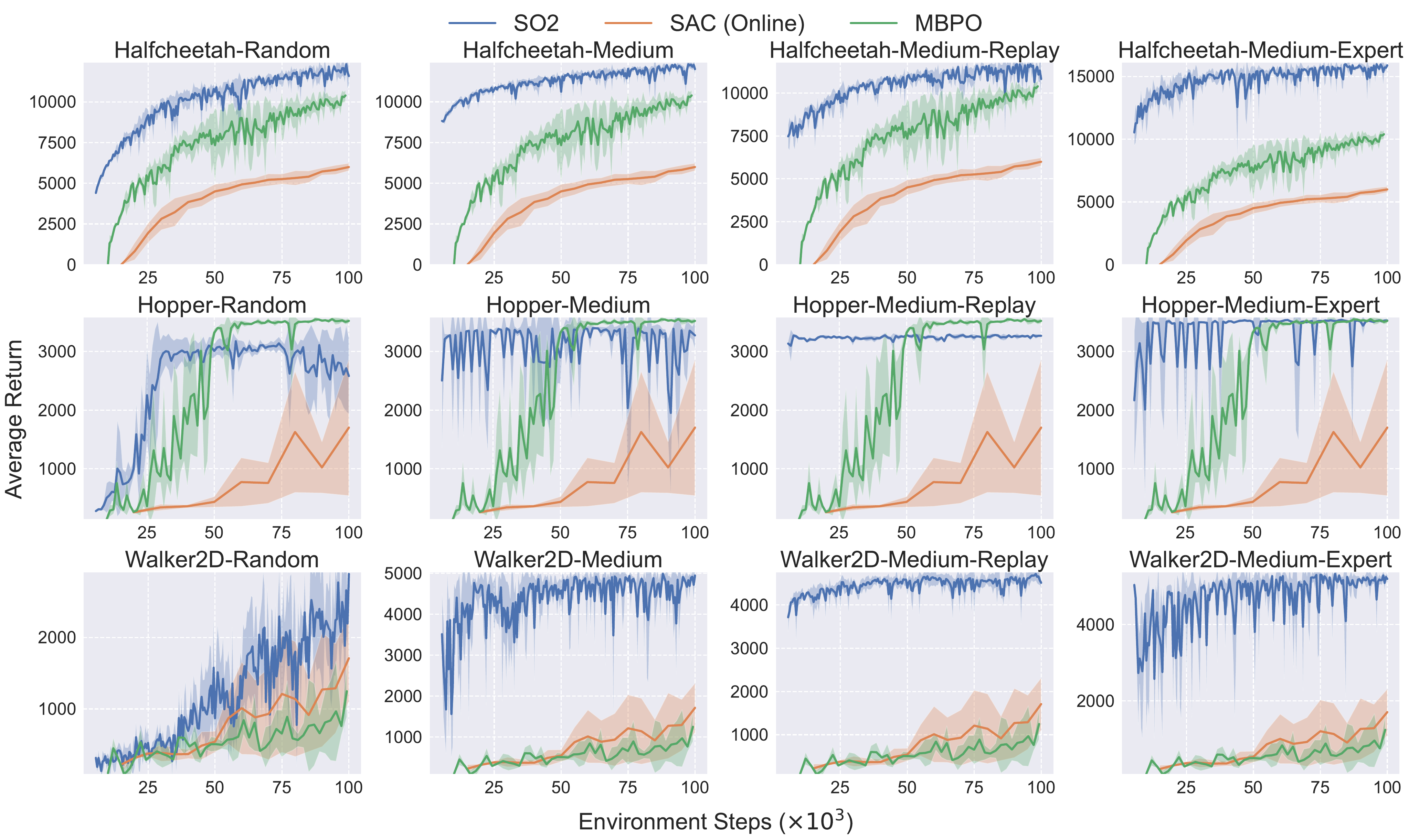}
    \caption{Learning curves comparing the performance of ours against SAC and MBPO. Learning curves are averaged over 4 seeds, and the shaded area represents the standard deviation across seeds. Our proposed method
outperforms SAC and MBPO by a large margin in terms of both sample efficiency and asymptotic performance with a small standard deviation in most environments.}
    \label{fig: appendix_mbpo}
\vspace{-2ex}
\end{figure*}

To better understand the advantage of offline-to-online RL on the sample efficiency, we compare our proposed offline-to-online finetuning RL method (SO2) with other deep RL algorithms, including online RL methods (off-policy SAC) and model-based RL (MBPO). In Figure~\ref{fig: appendix_mbpo}, our results show significant improvement in terms of sample efficiency compared with off-policy RL and model-based RL, even the model pretrained in the random dataset.

\paragraph{SAC.} an off-policy RL algorithm for continuous environments. We train a SAC agent from scratch without access to the offline dataset, highlighting the advantage of offline-to-online RL, compared to fully online RL, in terms of sample efficiency. SO2 outperforms SAC in all environments with pretrained models from different datasets.

\paragraph{MBPO.} a model-based RL algorithm with high sample efficiency. We train an MBPO agent from scratch without access to the offline dataset, which highlights the advantage of offline-to-online RL, compared to model-based RL, in terms of sample efficiency. We find that SO2 outperforms MBPO in most environments with pretrained models from different datasets, while underperforms MBPO in the Hopper environment due to the slow improvement. One hypothesis for this performance gap is due to the more pessimism in the Hopper environment compared with other environments (see Table~\ref{tab:edac_gym_params}).

\subsection{Combination SO2 with other methods}
\begin{figure*}[h]
    \vspace{-2ex}
    \centering
    \includegraphics[width=0.92\linewidth]{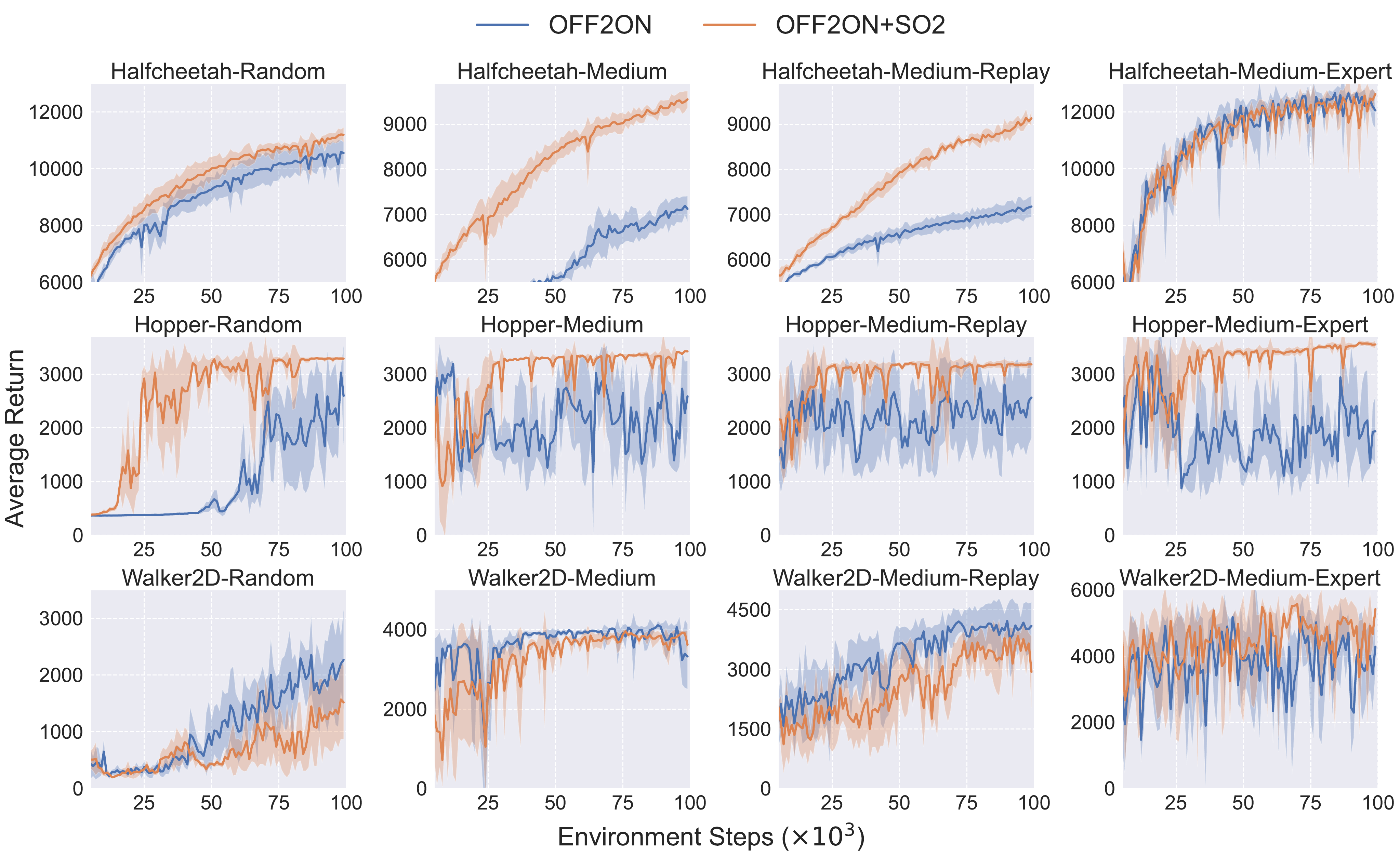}
    \caption{Learning curves comparing the performance of OFF2ON against OFF2ON+SO2 pretrained from D4RL datasets and finetuned with 100k environment steps. Learning curves are averaged over 4 seeds, and the shaded area represents the standard deviation across seeds. We observe that SO2 is compatible with methods like OFF2ON (Balanced Replay) which could encourage the use of near-on-policy samples from the ofﬂine dataset.}
    \label{fig: off2on}
\end{figure*}

To better understand the effectiveness of SO2 (\textit{i.g.}, PVU and $N_{upc}$) from a novel perspective, we combine SO2 with OFF2ON, the state-of-the-art offline-to-online finetuning RL algorithm. Figure~\ref{fig: off2on} compares OFF2ON-SO2 with OFF2ON, the state-of-the-art offline-to-online RL algorithm. We find that SO2 generally improves the final performance and training speed.
It also reveals that the proposed SO2 is compatible with methods like OFF2ON (Balanced Replay) which could encourage the use of near-on-policy samples from the ofﬂine dataset.

Moreover, we report the results of the combination of SO2 with OFF2ON and PEX~\cite{pex} in Table~\ref{tab: combination}. Our results demonstrate significant performance improvements when applying our proposed SO2 method to OFF2ON~\cite{off2on}, PEX~\cite{pex}, underscoring the effectiveness of our approach. Regarding the potential application of SO2 to different baseline RL algorithms, we believe that our method can be adapted to a wide range of RL algorithms, especially those that involve offline-to-online learning setups. In our experiments, we report results with 100K online samples on Mujoco environments, including halfcheetah, hopper, and walker2d, with varying difficulties ranging from random to medium, medium-replay, and medium-expert settings.

\begin{table}[h]
\centering
\begin{tabular}{l|l|l}
\toprule
       & 100k     & Algo. + SO2      \\
\midrule
OFF2ON & 75.89 ± 14.26 & 88.64 ± 4.33 \\
PEX    & 67.68 ± 8.83  & 74.26 ± 4.26 \\
\bottomrule
\end{tabular}
\caption{Compatibility of SO2. To validate the Compatibility of SO2 with other methods, we combine the SO2 with OFF2ON and PEX. We report results with 100K online samples on Mujoco environments, including halfcheetah, hopper, and walker2d, with varying difficulties ranging from random to medium, medium-replay, and medium-expert settings.}
\label{tab: combination}
\end{table}

\subsection{Only Increase Q Update Frequency}
\begin{figure*}[h]
    \centering
    \includegraphics[width=0.99\linewidth]{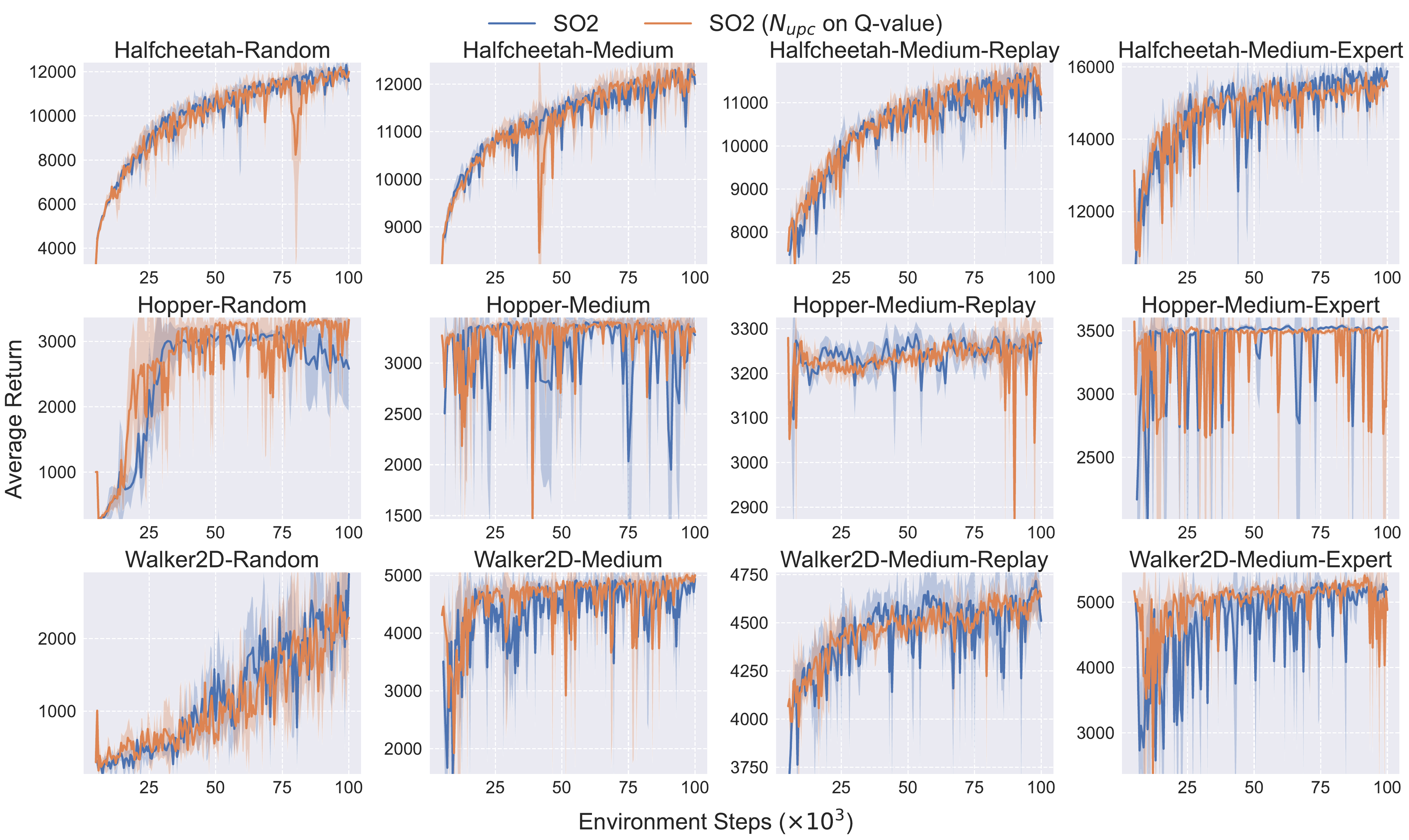}
    \caption{Learning curves comparing the performance of SO2 against SO2 with only increasing frequency of Q-value update. Learning curves are averaged over 4 seeds, and the shaded area represents the standard deviation across seeds. SO2 with only increasing the frequency of Q-value update performs comparatively to SO2 with increasing the frequency of both value updates and policy updates.}
    \label{fig: appendix_onlyq}
\vspace{-2ex}
\end{figure*}

A concern of SO2 is the $N_{upc}$ used for both value updates and policy updates, making it unclear which one has an effect on the learning speed and final performance.
In Figure~\ref{fig: appendix_onlyq}, we only increase the frequency of Q-value update.
We find that SO2 with only increasing frequency of Q-value update performs comparatively to SO2 with increasing frequency of both value updates and policy updates. However, considering it not elegant to separately modify the update frequency of Q-value update and policy update, we implement $N_{upc}$ by increasing the update frequency of Q-value update and policy update in the main body.

\subsection{Comparison with more baselines}
We additionally compare SO2 with more baselines on D4RL MuJoco benchmarks.
We reproduce all the methods by following the default hyperparameters listed in each paper. 
We report the normalized average return results in Table~\ref{tab: perf}.

\begin{table*}[h]
\centering
\resizebox{0.99\linewidth}{!}
{
\begin{tabular}{l|r|r|r|r|r|r}
\toprule
\multicolumn{2}{l|}{\textbf{Dataset}} & \multicolumn{1}{c|}{AWAC}		& \multicolumn{1}{c|}{Pretrained AWAC} 		& \multicolumn{1}{c|}{IQL}		& \multicolumn{1}{c|}{OFF2ON}		& \multicolumn{1}{c}{Ours} \\
\midrule
\multirow{3}{*}{\rotatebox{90}{Random}} 
& HalfCheetah & 6.5 $\pm$ 0.4 $\rightarrow$ 35.0 $\pm$ 1.1 	& 13.8 $\pm$ 1.6 $\rightarrow$ 34.2 $\pm$ 2.4 	& 13.4 $\pm$ 1.1 $\rightarrow$ 27.3 $\pm$ 6.0 	& 27.7 $\pm$ 1.0 $\rightarrow$ 87.2 $\pm$ 3.2 	& 37.7 $\pm$ 0.8 $\rightarrow$ 95.6 $\pm$ 7.2\\
& Hopper 	    & 6.7 $\pm$ 0.3 $\rightarrow$ 21.1 $\pm$ 11.2 	& 5.8 $\pm$ 2.3 $\rightarrow$ 7.1 $\pm$ 3.9 	& 7.6 $\pm$ 0.2 $\rightarrow$ 9.3 $\pm$ 0.8 	& 10.5 $\pm$ 0.0 $\rightarrow$ 80.4 $\pm$ 28.5 	& 9.2 $\pm$ 0.7 $\rightarrow$ 79.9 $\pm$ 13.9\\
& Walker2d 	& 5.9 $\pm$ 0.4 $\rightarrow$ 6.3 $\pm$ 1.4 	& 0.2 $\pm$ 0.6 $\rightarrow$ 2.4 $\pm$ 1.4 	& 6.8 $\pm$ 0.7 $\rightarrow$ 9.9 $\pm$ 1.6 	& 10.3 $\pm$ 3.4 $\rightarrow$ 49.4 $\pm$ 19.1 	& 6.9 $\pm$ 2.0 $\rightarrow$ 62.9 $\pm$ 13.4\\

\midrule

\multirow{3}{*}{\rotatebox{90}{\begin{tabular}[c]{@{}c@{}}Medium\\ Replay\end{tabular}}}
& HalfCheetah & 40.5 $\pm$ 1.2 $\rightarrow$ 41.2 $\pm$ 1.2 	& 42.1 $\pm$ 0.6 $\rightarrow$ 42.5 $\pm$ 1.1 	& 42.7 $\pm$ 4.1 $\rightarrow$ 36.7 $\pm$ 16.7 	& 42.1 $\pm$ 2.5 $\rightarrow$ 60.0 $\pm$ 1.9 	& 62.5 $\pm$ 5.4 $\rightarrow$ 89.4 $\pm$ 5.8\\
& Hopper 	    & 37.7 $\pm$ 15.8 $\rightarrow$ 60.1 $\pm$ 20.9 	& 39.5 $\pm$ 6.3 $\rightarrow$ 53.3 $\pm$ 8.8 	& 75.8 $\pm$ 23.8 $\rightarrow$ 68.5 $\pm$ 20.7 	& 28.2 $\pm$ 0.6 $\rightarrow$ 79.5 $\pm$ 23.3 	& 97.0 $\pm$ 1.2 $\rightarrow$ 101.0 $\pm$ 0.5\\
& Walker2d 	& 24.5 $\pm$ 13.3 $\rightarrow$ 79.8 $\pm$ 5.6 	& 73.7 $\pm$ 5.1 $\rightarrow$ 77.2 $\pm$ 15.0 	& 75.6 $\pm$ 20.0 $\rightarrow$ 64.9 $\pm$ 17.3 	& 17.7 $\pm$ 4.3 $\rightarrow$ 89.2 $\pm$ 13.0 	& 80.9 $\pm$ 7.2 $\rightarrow$ 98.2 $\pm$ 12.8\\
\midrule

\multirow{3}{*}{\rotatebox{90}{Medium}}
& HalfCheetah & 43.0 $\pm$ 1.3 $\rightarrow$ 42.4 $\pm$ 1.1 	& 44.2 $\pm$ 0.5 $\rightarrow$ 44.4 $\pm$ 0.9 	& 46.7 $\pm$ 0.4 $\rightarrow$ 46.5 $\pm$ 0.5 	& 39.3 $\pm$ 0.3 $\rightarrow$ 59.6 $\pm$ 2.3 	& 73.3 $\pm$ 0.5 $\rightarrow$ 98.9 $\pm$ 2.4\\
& Hopper 	    & 57.8 $\pm$ 14.2 $\rightarrow$ 55.1 $\pm$ 7.4 	& 58.0 $\pm$ 2.0 $\rightarrow$ 54.7 $\pm$ 6.5 	& 73.4 $\pm$ 18.6 $\rightarrow$ 61.9 $\pm$ 12.6 	& 67.5 $\pm$ 5.0 $\rightarrow$ 80.2 $\pm$ 20.3 	& 77.6 $\pm$ 35.3 $\rightarrow$ 101.2 $\pm$ 2.9\\
& Walker2d 	& 35.9 $\pm$ 32.2 $\rightarrow$ 72.1 $\pm$ 13.9 	& 66.8 $\pm$ 20.7 $\rightarrow$ 77.3 $\pm$ 8.4 	& 84.7 $\pm$ 3.0 $\rightarrow$ 78.8 $\pm$ 10.8 	& 66.2 $\pm$ 8.6 $\rightarrow$ 72.4 $\pm$ 17.9 	& 76.4 $\pm$ 21.6 $\rightarrow$ 107.6 $\pm$ 10.9\\

\midrule
\multirow{3}{*}{\rotatebox{90}{\begin{tabular}[c]{@{}c@{}}Medium\\ Expert\end{tabular}}}
& HalfCheetah & 65.2 $\pm$ 25.1 $\rightarrow$ 93.0 $\pm$ 0.9 	& 92.9 $\pm$ 0.4 $\rightarrow$ 93.2 $\pm$ 1.1 	& 88.2 $\pm$ 4.3 $\rightarrow$ 59.6 $\pm$ 33.0 	& 56.7 $\pm$ 7.7 $\rightarrow$ 99.3 $\pm$ 5.1 	& 87.1 $\pm$ 8.2 $\rightarrow$ 130.2 $\pm$ 2.1\\
& Hopper 	    & 55.5 $\pm$ 11.0 $\rightarrow$ 81.3 $\pm$ 25.2 	& 81.9 $\pm$ 25.1 $\rightarrow$ 76.6 $\pm$ 22.6 	& 52.8 $\pm$ 25.5 $\rightarrow$ 65.0 $\pm$ 42.9 	& 112.0 $\pm$ 0.3 $\rightarrow$ 60.2 $\pm$ 19.9 	& 67.2 $\pm$ 36.2 $\rightarrow$ 108.8 $\pm$ 0.7\\
& Walker2d 	& 108.3 $\pm$ 0.6 $\rightarrow$ 108.7 $\pm$ 0.2 	& 109.0 $\pm$ 0.5 $\rightarrow$ 109.3 $\pm$ 0.3 	& 109.1 $\pm$ 0.7 $\rightarrow$ 110.4 $\pm$ 1.4 	& 102.1 $\pm$ 7.7 $\rightarrow$ 93.3 $\pm$ 16.8 	& 109.5 $\pm$ 1.8 $\rightarrow$ 112.9 $\pm$ 23.1 \\

\midrule
\multicolumn{1}{l}{} 
&\textbf{Average} & 40.6 $\pm$ 9.7 $\rightarrow$ 58.0	$\pm$ 7.5 
                & 52.3	$\pm$ 5.5 $\rightarrow$ 56.0 $\pm$	6.0
                & 56.6	$\pm$ 8.5 $\rightarrow$ 53.2    $\pm$ 13.7 
                
                &48.4 $\pm$	3.5 $\rightarrow$  75.9 $\pm$	14.3
                & 76.1 $\pm$ 2.4 $\rightarrow$ 98.9 $\pm$ 5.4 \\
\bottomrule
\end{tabular}
}
\caption{Normalized average returns on D4RL Gym tasks, averaged over 4 random seeds. $\pm$ captures the standard deviation over seeds. SO2 effectively and efficiently achieves competitive performance improvement and remarkably improves performance compared to the state-of-the-art OFF2ON with a small standard deviation, despite being much simpler to implement, exhibiting that SO2 generally improves the final performance and training stability.}
\label{tab: perf}
\end{table*}



\end{document}